\documentclass[runningheads]{llncs}
\usepackage[T1]{fontenc}
\usepackage{amsmath}
\usepackage{amssymb} 
\usepackage{graphicx}
\usepackage{array}
\usepackage{booktabs}  
\usepackage{colortbl}
\usepackage{arydshln}
\usepackage[dvipsnames]{xcolor}
\usepackage{algorithm}  
\usepackage{bbding}
\usepackage{algpseudocode}  

\usepackage{hyperref}
\usepackage{multirow}
\begin{document}
\title{RAPID: Retrieval-Augmented Parallel Inference Drafting for Text-Based Video Event Retrieval}

\titlerunning{RAPID: A Novel Approach for Text-Based Video Event Retrieval}
\author{Long Nguyen\inst{\star}\orcidID{0009-0008-7488-4714} \and Huy Nguyen\inst{\star} \and  Bao Khuu \and Huy Luu \and Huy Le \and Tuan Nguyen \and Tho Quan\orcidID{0000-0003-0467-6254}}

\renewcommand{\thefootnote}{\fnsymbol{footnote}}
\footnotetext[1]{The two authors contributed equally to this work.}
\authorrunning{Long Nguyen, Huy Nguyen, et al.}
\institute{URA Research Group, Faculty of Computer Science and Engineering, Ho Chi Minh City University of Technology (HCMUT), Ho Chi Minh City, Vietnam \and Vietnam National University Ho Chi Minh City,  Ho Chi Minh City, Vietnam}
\maketitle              % typeset the header of the contribution
\begin{abstract}
Retrieving events from videos using text queries has become increasingly challenging due to the rapid growth of multimedia content. Existing methods for text-based video event retrieval often focus heavily on object-level descriptions, overlooking the crucial role of contextual information. This limitation is especially apparent when queries lack sufficient context, such as missing location details or ambiguous background elements. To address these challenges, we propose a novel system called RAPID (Retrieval-Augmented Parallel Inference Drafting), which leverages advancements in Large Language Models (LLMs) and prompt-based learning to semantically correct and enrich user queries with relevant contextual information. These enriched queries are then processed through parallel retrieval, followed by an evaluation step to select the most relevant results based on their alignment with the original query. Through extensive experiments on our custom-developed dataset, we demonstrate that RAPID significantly outperforms traditional retrieval methods, particularly for contextually incomplete queries. Our system was validated for both speed and accuracy through participation in the Ho Chi Minh City AI Challenge 2024, where it successfully retrieved events from over 300 hours of video. Further evaluation comparing RAPID with the baseline proposed by the competition organizers demonstrated its superior effectiveness, highlighting the strength and robustness of our approach.

\keywords{Text-Based Video Event Retrieval \and Contextual Query Enrichment  \and Parallel Inference \and Multimedia and Multimodal Retrieval.}
\end{abstract}

\section{Introduction}

The rapid advancement of \textit{Artificial Intelligence} (AI), particularly in the field of Computer Vision, has driven significant breakthroughs in video event retrieval. Initially, research in this area focused on applications in Intelligent Transport Systems \cite{954548}, but it has since expanded to include surveillance, security, and multimedia content analysis \cite{10.1007/978-3-642-17691-3_11}. With the exponential growth of video data across online platforms, the ability to efficiently retrieve specific events from vast video datasets has emerged as a critical research challenge \cite{7410875}. Developing more efficient and accurate retrieval methods is essential to managing the overwhelming volume of media content generated daily and improving retrieval accuracy \cite{gabeur2020multi}. 

\renewcommand{\thefootnote}{\arabic{footnote}}
To address the growing demand for efficient video retrieval solutions, several international competitions, such as the \textit{Lifelog Search Challenge} (LSC) \cite{10.1145/3652583.3658891} and the \textit{Video Browser Showdown} (VBS) \cite{8877397}, have been organized to foster the development of fast and effective video search methods. Similarly, in Vietnam, the \textit{Ho Chi Minh City AI Challenge 2024}\footnote{\url{http://aichallenge.hochiminhcity.gov.vn/}} was established to encourage innovation in video event retrieval. In this competition, participants were tasked with retrieving events from over 300 hours of 1,500 news videos using a series of sequential text-based queries or short visual clips depicting the events. Notably, these queries often began with broad, context-poor descriptions, and many video segments lacked clear contextual cues, such as background details, making accurate retrieval a significant challenge.

Various approaches to video event retrieval have been explored over the years, leveraging a diverse range of input modalities. These include image frames \cite{895972}, object tags \cite{AMIR2004216}, event labels \cite{10.1145/2578726.2578793}, and even audio data \cite{Jin2012EventbasedVR}. More advanced methods use high-level sketches to visually represent events \cite{Zhang2016HighlevelRS}. Recent progress in Natural Language Processing, especially with the development of multimodal models like \textit{Contrastive Language-Image Pre-training} (CLIP) \cite{pmlr-v139-radford21a}, has significantly advanced text-based video retrieval. These models map text queries and video frames into a shared embedding space using contrastive learning, enabling more intuitive and effective retrieval \cite{9954207,9523150}. However, the effectiveness of text-based retrieval systems heavily depends on the completeness of the query. Queries that lack sufficient context often lead to reduced retrieval accuracy \cite{9354593}, underscoring the need for enriched contextual information \cite{10.1145/3589335.3651942}. Moreover, certain events are complex and difficult to describe in detail, further complicating the retrieval process.

To address these challenges, we propose \textit{Retrieval-Augmented Parallel Inference Drafting} (RAPID), a novel method designed to enhance text-based video event retrieval, particularly in scenarios where queries lack sufficient context and where frames are difficult to describe. Unlike traditional methods that rely solely on single, unrefined queries, RAPID generates multiple \textit{augmented queries} by enriching the original with additional contextual details, such as location or event-specific information. These \textit{drafts} are created using \textit{Large Language Models} (LLMs) combined with prompt-based learning techniques \cite{NEURIPS2020_1457c0d6}. By performing \textit{parallel inference}, RAPID retrieves the most relevant keyframes, which are subsequently re-evaluated for alignment with the original query. Additionally, we developed a user-friendly interface to support practical, real-world application of the system.

% \begin{figure}
%     \centering
%     \includegraphics[width=\linewidth]{IMAGES/pipeline.pdf}
%     \caption{Overview of the RAPID framework.}
%     \label{fig:pipeline}
% \end{figure}

We evaluated RAPID using a dataset derived from 300 hours of news videos provided by the challenge. Our experimental results demonstrate that enriching queries with contextual information, such as location, significantly improves retrieval accuracy and efficiency. Furthermore, RAPID outperformed the baseline retrieval pipeline provided by the competition organizers, showcasing its ability to handle real-world video retrieval challenges, especially when dealing with incomplete or context-poor queries.

In summary, our key contributions are as follows.
\begin{itemize}
    \item We developed the RAPID system, which leverages parallel processing of multiple augmented queries to significantly improve video event retrieval performance.
    \item We demonstrated the effectiveness of RAPID in handling incomplete or context-poor queries, and compared its performance against the baseline provided by the competition, showing notable improvements.
    \item We successfully applied RAPID in a real-world scenario, achieving excellent results at the Ho Chi Minh City AI Challenge 2024, highlighting its accuracy and speed.
\end{itemize}

\section{Related Work}

\subsection{Text-based Video Event Retrieval Methods}

Various video event retrieval methods have evolved from traditional text-based queries that match against metadata such as titles and tags \cite{AMIR2004216,5729374,10.1145/2578726.2578793}. These methods often fail when metadata is incomplete or when users need to locate complex events that are difficult to describe using simple tags or titles. To address these limitations, visual-based retrieval techniques, such as content-based image retrieval \cite{895972} and query-by-high-level sketches \cite{10.1007/978-3-031-27077-2_53,Zhang2016HighlevelRS}, were introduced to better analyze the visual content of videos. With the notable success of multi-modal models like CLIP \cite{pmlr-v139-radford21a}, baseline systems that retrieve both text and image frames from videos in a shared vector space have become increasingly effective \cite{9523150,10.1145/3628797.3629011}. These advancements have enabled more intuitive and accurate retrieval processes, especially when handling large-scale datasets. However, despite the improved performance of modern retrieval systems, challenges remain, particularly in aligning different modalities. This process requires substantial computational resources and poses scalability challenges for real-time applications \cite{Tian_2024_AAAI}. Moreover, these methods heavily rely on the accuracy and completeness of text queries as the primary alignment mechanism. When event descriptions are vague or imprecise, system effectiveness is reduced. Research has shown that including contextual components in queries, such as location information, can significantly enhance model performance, as highlighted in \cite{9354593,10.1145/3589335.3651942}.

\subsection{Augmented Query Strategies for Video Event Retrieval}

The rapid evolution of LLMs, combined with the power of prompt-based learning across various tasks \cite{NEURIPS2020_1457c0d6}, has driven significant interest in developing more sophisticated systems for information retrieval \cite{AI202380}. By improving the semantic embedding of queries, LLMs can enhance retrieval systems by providing richer, more context-aware representations of the user's search intent. This is particularly useful when the original query lacks detailed context or contains ambiguities, as the augmented query versions generated by LLMs can fill in the missing information, thereby improving the overall precision and recall in video retrieval systems \cite{10.1145/3604915.3610639}. LLMs can also suggest query completions or augmentations, as demonstrated in fashion recommendation systems, where user history is leveraged to improve query suggestions \cite{10678412}. Similarly, research \cite{AI202380} has shown that LLMs can enhance the semantic content of queries in retrieval systems, making them more context-aware and better aligned with the user’s intent. Specifically, in the domain of video event retrieval, studies \cite{wu2024llm,10.1145/3652583.3658032} have demonstrated the effectiveness of rewriting queries logically to automatically incorporate additional context, thereby significantly improving retrieval performance. However, despite the benefits, enriching queries with LLM-generated augmentations also carries risks, such as introducing irrelevant or excessive information, which can decrease efficiency and accuracy. Moreover, some events within video datasets are inherently complex, making it difficult to accurately describe their surrounding context through visual inspection alone. This further complicates the creation of detailed and context-rich queries necessary for effective video event retrieval. 

\section{RAPID: A Novel Approach for Text-Based Video Event Retrieval}

\subsection{Motivation}
Previous studies have demonstrated the potential of augmented retrieval methods for video event retrieval using text-based queries. Building on the concept of \textit{EmotionPrompt}, which highlights how enriching prompts with additional elements, such as positive emotions, can improve LLM performance \cite{Li2023LargeLM}, we adapt this approach by focusing on context-specific elements, particularly location information, to enhance query relevance. Additionally, inspired by Google’s recent \textit{Speculative RAG} research \cite{Wang2024SpeculativeRE}, which explores parallel query processing across multiple drafts, we propose \textit{RAPID}, a novel method for text-based event retrieval. RAPID augments the original query by generating several variations enriched with contextual details, such as location information, and processes these queries in parallel to improve both retrieval accuracy and speed. The formal methodology behind RAPID is described in Subsection \ref{sec:maths}.

\subsection{Methodology}
\label{sec:maths}
The notation used throughout this paper is summarized in Table \ref{tab:notation}.
\setlength{\tabcolsep}{3pt} % Adjust column separation to increase spacing between columns
\renewcommand{\arraystretch}{1.2}
\medskip
\begin{table}[!hbt]
    \centering
    \caption{Notation}
    \label{tab:notation}
    \resizebox{\textwidth}{!}{%
    \begin{tabular}{>{\raggedright\arraybackslash}p{3.3cm} p{8.7cm}}
        \toprule
        \textbf{Symbol} & \textbf{Description} \\
        \midrule
        \(\mathbb{T}\) & Text space \\
        \(\mathbb{I}\) & Image space \\
        \(\mathcal{M}_{m \times n}\) & Matrix with \(m\) rows and \(n\) columns \\
        \(\mathcal{L}: \mathbb{T} \to \mathbb{T}\) & Function that transforms the original query into augmented versions by adding varying contextual details, particularly location information, leveraging a \textsf{Large Language Model} \\
        \(\mathcal{E}: \mathbb{T} \cup \mathbb{I} \to \mathbb{R}^d\) & Multi-modal embedding function that maps both text and images to a shared \(d\)-dimensional vector space via a \textsf{Multi-modal Embedding Model} \\
        \(\text{Idx}_k: \mathcal{M}_{1 \times n} \to \mathcal{M}_{1 \times k}\) & Function that retrieves the indices of the top-\(k\) highest values from a vector \(\mathcal{M}\) with \(n\) values \\
        \bottomrule
    \end{tabular}
    }
\end{table}

\noindent \textbf{Task Definition} \quad 
We define the task of text-based event retrieval, particularly in the context of the Ho Chi Minh City AI Challenge, as identifying a frame \( F_c \in \mathcal{F} \), where \( \mathcal{F} \) denotes the set of frames relevant to the described event, from a video corpus \(\mathcal{V} = \{V_1, V_2, .., V_p\}\) totaling approximately 300 hours of footage. Given a text query \( Q_0 \), the objective is to retrieve a frame \( F_c \) such that \( F_c \in \mathcal{F} \) and best matches the content of \( Q_0 \). For visual queries, represented by short video clips \( V_0 \), we employ a strategy of converting these videos into descriptive text queries \( Q_0 \). As a result, both textual and visual queries are handled under a unified text-based retrieval framework.

\medskip
\noindent \textbf{Data Preprocessing} \quad
Each video \(V_i \in \mathcal{V}\subset \mathbb{I}\) consists of image frames, where a video recorded at \(x\) \textit{frames per second} (fps) produces \(x\) frames for every second of playback. To reduce computational complexity, we extract \textit{keyframes}, which are representative frames containing the key visual information for each segment \cite{10203658}. For this, we employ \textit{TransNet V2} \cite{Soucek2020TransNetVA}, an effective deep learning architecture optimized for fast shot transition detection. For each video \(V_i\), we generate a set of scenes \(\mathcal{S}_i = \{S_{i_1}, S_{i_2}, .., S_{i_{s_i}}\}\), where each scene \(S_{i_j}\) consists of frames \(\{S_{i_{j_1}}, S_{i_{j_2}}, .., S_{i_{j_e}}\} \subset \mathbb{I}\). From each scene \(j\), we select three keyframes: the first, middle, and last frames, as described in Equation \ref{eq:keyframes}.
\begin{equation}
S_{i_{j_\text{selected}}} = \{S_{i_{j_1}}, S_{i_{j_{\lfloor \frac{e-1}{2} \rfloor}}}, S_{i_{j_e}}\}.
\label{eq:keyframes}
\end{equation}
These keyframes are aggregated into the set \(\mathcal{F} \subset \mathbb{I}\), as described in Equation \ref{eq:keyframe_set}.
\begin{equation}
\mathcal{F} = \bigcup_{i=1}^{p} \bigcup_{j=1}^{s_i} S_{i_{j_\text{selected}}},
\label{eq:keyframe_set}
\end{equation}
where \(p\) is the total number of videos, and \(s_i\) is the number of scenes in video \(i\).

\medskip
\noindent \textbf{RAPID Inference} \quad
Let \(Q_0\) represent the initial query, which may lack sufficient context.

The function \(\mathcal{L}(Q_0)\) generates a set of \textit{augmented queries}, denoted as \(\mathcal{Q} = \{Q_1, Q_2, .., Q_n\}\). Each augmented query \(Q_i\) is then embedded as \(\mathbf{Q} = \mathcal{E}(\mathcal{Q}) = \{\mathbf{q}_1, \mathbf{q}_2, .., \mathbf{q}_n\}\).

The keyframes obtained from preprocessing, \(\mathcal{F} = \{F_1, F_2, .., F_m\}\), where \(\displaystyle m = \sum_{i=1}^{p} s_i \times 3\), are also embedded, yielding \(\mathbf{F} = \mathcal{E}(\mathcal{F}) = \{\mathbf{f}_1, \mathbf{f_2}, .., \mathbf{f}_m\}\).

Both the augmented \textit{drafts} \(\mathcal{Q}\) and the set of keyframes \(\mathcal{F}\) are embedded in the shared vector space \(\mathbb{R}^d\). To enable \textit{parallel retrieval}, we compute the cosine similarity between the query embeddings \(\mathbf{q}_i\) and the keyframe embeddings \(\mathbf{f}_j\), as shown in Equation \ref{eq:similarity}.
\begin{equation}
\mathcal{P}=\text{sim}(\mathbf{Q}, \mathbf{F}) = \frac{\mathbf{Q} \cdot \mathbf{F}^\top}{\|\mathbf{Q}\| \|\mathbf{F}\|}  \in \mathcal{M}_{n \times m}.
\label{eq:similarity}
\end{equation}

We then retrieve the top-\(k\) most similar frames for each draft \(Q_i\), as shown in Equation \ref{eq:topk_frames}.
\begin{equation}
\mathcal{C}_i = \text{Idx}_k(\mathcal{P}_{i,:}), \quad \forall i = 1, 2, .., n.
\label{eq:topk_frames}
\end{equation}
This produces a matrix \(\mathcal{C} \in \mathcal{M}_{n \times k}\), where each row contains the indices of the top-\(k\) frames for each augmented query. These selected frames are flattened into a set \(\mathcal{K} = \{F_1, F_2, .., F_w\}\), where \(w = n \times k\).

Finally, these frames are re-evaluated based on their similarity to the original query \(Q_0\), as shown in Equation \ref{eq:final_topk}.
\begin{equation}
\mathcal{C}_{\text{final}} = \text{Idx}_K(\text{sim}(\mathcal{E}(Q_0), \mathcal{E}(\mathcal{K}))) \in \mathcal{M}_{1 \times K},
\label{eq:final_topk}
\end{equation}
where \(\mathcal{C}_{\text{final}}\) represents the indices of the top-\(K\) frames that best match the original query \(Q_0\). 

The final set of selected frames is presented to the user for evaluation, allowing them to verify whether the retrieved frames not only match the context of the query but also align with the specific information they are searching for.

\medskip
\noindent \textbf{Multi-modal Embedding and Prompting Techniques} \quad RAPID integrates multi-modal embeddings and advanced prompting strategies to optimize the alignment between text and visual data, enhancing retrieval performance.

A key property of the multi-modal embedding function \(\mathcal{E}(\cdot)\), implemented using a \textsf{Multi-modal Embedding Model}, is that for any image-text pair \((I_i, T_i)\), where \(T_i\) accurately describes \(I_i\), the cosine similarity between their embeddings, \(\cos \theta (\mathcal{E}(I_i), \mathcal{E}(T_i))\), approaches \(1\). This property ensures that the more semantically aligned the text query is with the image, the closer their embeddings are in the shared vector space \(\mathbb{R}^d\), forming the backbone of the RAPID approach.

To enhance the drafting process, denoted as \(\mathcal{L}(\cdot)\), RAPID utilizes a \textsf{Large Language Model} combined with few-shot prompting \cite{NEURIPS2020_1457c0d6}, specifically employing the \textit{Chain-of-Thought} (CoT) reasoning technique \cite{wei2022chain}. The prompt provided to the LLM for augmenting the query \(q_0\) is formally described in Equation \ref{eq:prompt}.
\begin{align}
\label{eq:prompt}
    \text{prompt}(q_0) &= \sum_{i=1}^{n} \langle q_i, \text{CoT}(q_i), \{q_{\text{augmented}_i^j}\}_j \rangle + q_0, \\
    \text{prompt}(q_0) &\to \{q_{\text{augmented}_0^j}\}_j \notag,
\end{align}
where \(q_1, q_2, .., q_n\) represent example queries, and each \(q_i\) is associated with a chain of reasoning \(\text{CoT}(q_i)\), which includes logical reasoning steps to guide the LLM, such as identifying key content and verifying contextual information. This reasoning process generates a set of augmented queries \(\{q_{\text{augmented}_i^j}\}_j\). Once these examples are processed, the LLM receives the target query \(q_0\) and applies the learned patterns to produce a set of augmented queries \(\{q_{\text{augmented}_0^j}\}_j\).

\subsection{User Inference Design and Interaction}

\begin{figure}[!ht]
    \centering
    \label{fig:monk}
    \includegraphics[width=\textwidth]{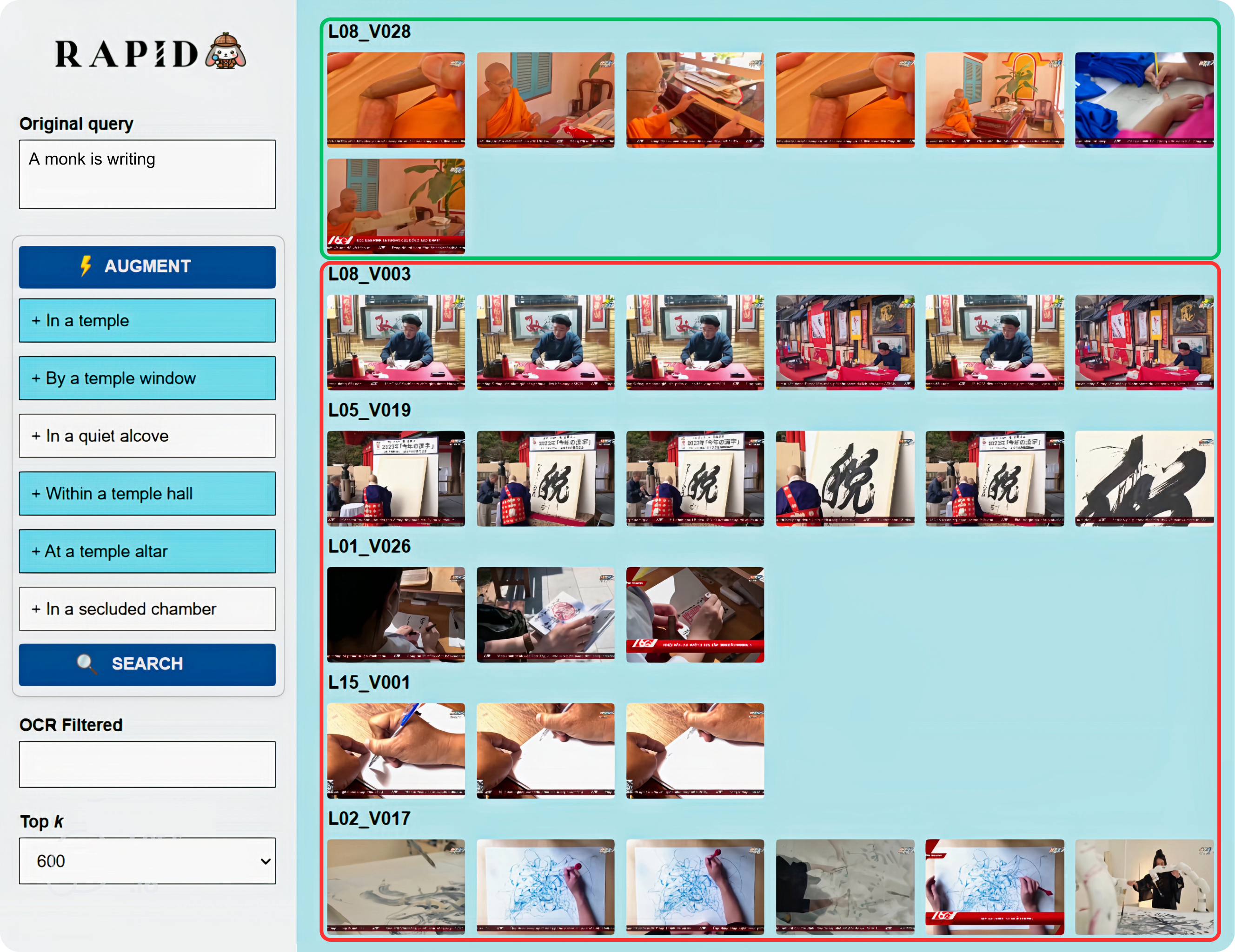}
    \caption{An illustration of RAPID's UI for the textual query \( Q_0 \): \textit{A monk is writing}, where \( n = 4 \) augmented queries are selected from \( N = 6 \) generated drafts, and the parameter \( K = 600 \) specifies the number of final keyframes. The relevant result, highlighted in green, is displayed among the top-ranked keyframes.}
    \label{fig:enter-label}
\end{figure}

To enable the deployment of RAPID at the Ho Chi Minh City AI Challenge 2024, we developed an intuitive \textit{User Interface} (UI) tailored to RAPID’s operational workflow, as illustrated in Figure \ref{fig:monk}. The process begins when the user inputs the initial query \( Q_0 \) in the \texttt{Original Query} field and presses the \texttt{Augment} button. This triggers the system to generate \( N \) augmented queries \( \{Q_1, Q_2, .., Q_N\} \). The user then selects \( n \) suitable augmented queries and initiates the inference process by pressing the \texttt{Search} button, following the pipeline described in Subsection \ref{sec:maths}.

% After RAPID inference, we obtain a ranked list of the top \( K \) keyframes, denoted as \( \mathcal{L}_K = \{F_1, F_2, .., F_K\} \). Displaying these keyframes sequentially could result in a cluttered view, as multiple keyframes may originate from the same video \( V_i \). To improve clarity and organization, we group the keyframes in \( \mathcal{L}_K \) by their source videos, forming clusters \( \{C_{V_1}, C_{V_2}, .., C_{V_m}\} \), where each \( C_{V_i} \subset \mathcal{L}_K \) contains all keyframes from video \( V_i \in \mathcal{V} \). Within each cluster, the keyframes are ordered according to their appearance in \( \mathcal{L}_K \), preserving their relative ranking. These clusters are then displayed on the UI in the order of the first appearance of their keyframes in \( \mathcal{L}_K \), ensuring that clusters with higher-ranked keyframes appear first. This structured approach provides an intuitive display, allowing users to view related frames grouped by video while maintaining the overall relevance order.

Additionally, we incorporated a filtering feature using an \textit{Optical Character Recognition} (OCR) model, \( \mathcal{O}(\cdot) \). This feature enables the system to apply a filter \( \mathcal{O}(\mathcal{L}) = \mathcal{L}_K' \), where \( \mathcal{L}_K' \) represents the subset of keyframes containing the desired text. This capability is particularly valuable when searching for event frames that include visible text, thereby enhancing the system’s effectiveness in retrieving context-specific keyframes.

\begin{figure}[!ht]
    \centering
    \label{fig:monk1}
    \includegraphics[width=0.7\textwidth]{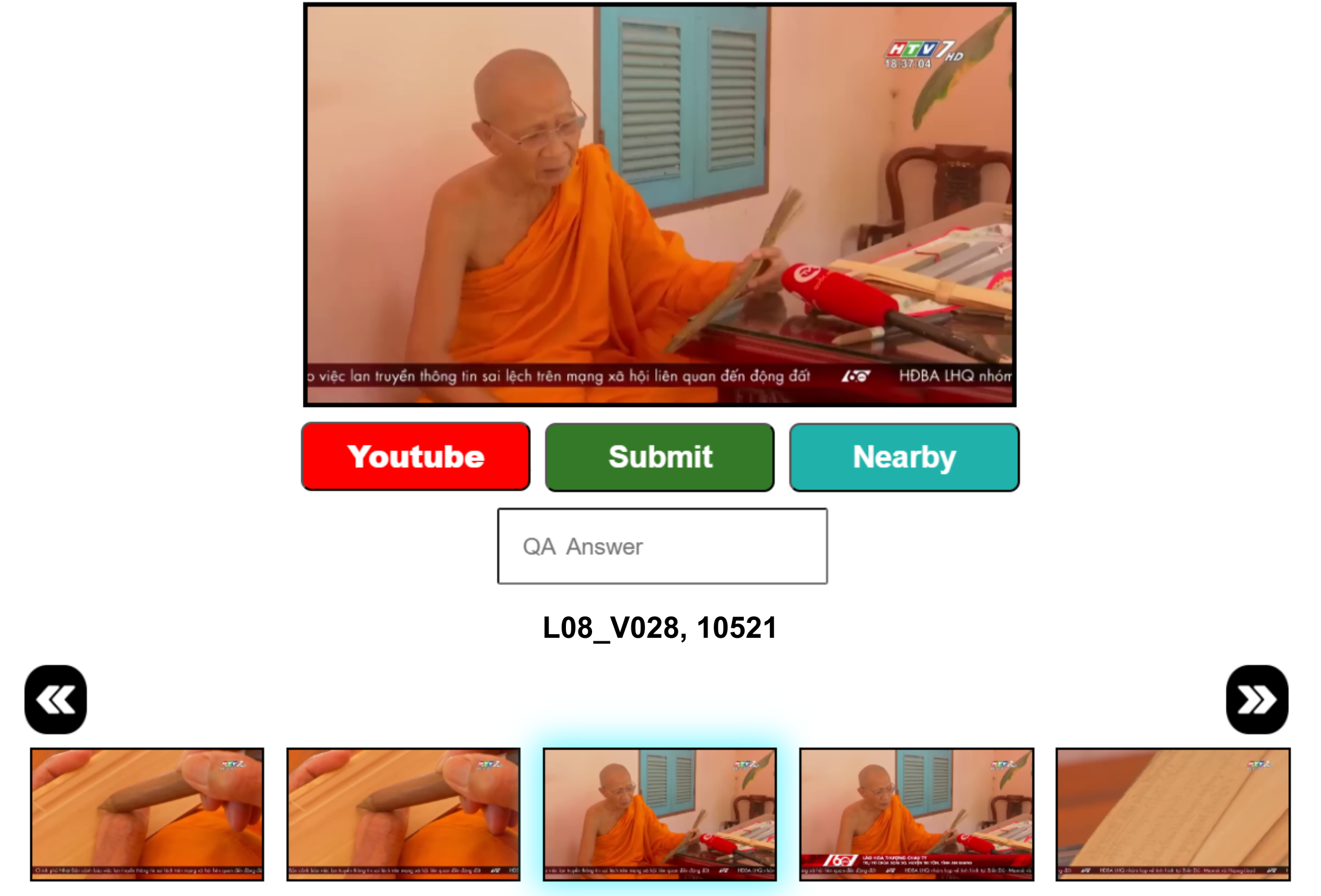}
    \caption{The user can review frames adjacent to the selected keyframe to check for accuracy before pressing \texttt{Submit} and can view the complete video containing it by pressing the \texttt{Youtube} button.}
\end{figure}

Another key feature, depicted in Figure \ref{fig:monk1}, allows users to review a sequence of \( \pi \) keyframes before and after a selected frame \( F_c \), as defined in Equation \ref{eq:related}.
\begin{equation}\label{eq:related}
    \left\{F_{c-\pi}, .., F_{c-1}, F_c, F_{c+1}, .., F_{c+\pi}\right\}.
\end{equation}
In this context, assume \( F_c \) is the target frame to be identified, which corresponds to the description \( Q_0 \) provided by the organizers. However, visually similar frames such as \( F'_c \in V_i \) and \( F_c \in V_j \), where \( i \neq j \) and \( F'_c \approx F_c \), can cause confusion between events. To address this issue, the organizers provide additional queries \( Q'_0 \) that include detailed information about the event, specifically describing the frames surrounding \( F_c \). This additional context helps users accurately identify the correct frame and avoid confusion with visually similar frames like \( F'_c \). This feature is therefore essential for distinguishing between events.

\section{Experimentation} 
We conducted two sets of experiments to evaluate the effectiveness of the RAPID system: one comparing retrieval performance between location-augmented queries and non-augmented queries, and the other assessing the overall performance of RAPID against the baseline system provided by the Ho Chi Minh City AI Challenge 2024, referred to as the \textit{AIC'24 Baseline}.

\subsection{Experimental Setup}

\noindent \textbf{Dataset} \quad We utilized a dataset consisting of 300 hours of news video footage provided by the challenge organizers. From this dataset, keyframes were extracted, and we selected keyframes corresponding to specific events, categorized into two types: \textit{Type 1}, with clearly identifiable locations, and \textit{Type 2}, where the location is less discernible. For each set of event keyframes, descriptive queries were generated by focusing on prominent visual elements such as character actions and visible objects. The final dataset consists of 500 records, each containing a \verb|query| (a textual description of the frame) and the corresponding \verb|keyframes| (range of frame IDs). Of these, 84\% are classified as Type 1, while the remaining 16\% fall under Type 2.

\medskip 
\noindent \textbf{Evaluation Metrics} \quad To evaluate the system's performance, we employed well-established information retrieval metrics, including \textit{Mean Reciprocal Rank} (MRR), \textit{Precision@$k$} (P@$k$), and \textit{Recall@$k$} (R@$k$) \cite{10.1145/3460231.3478848}. MRR measures how early the correct frame is retrieved, Precision@$k$ evaluates the proportion of relevant results in the top-$k$ retrieved frames, and Recall@$k$ assesses the system's ability to retrieve all relevant frames within the top-$k$. These metrics are highly suitable for evaluating text-based video event retrieval systems, as they emphasize both the accuracy and coverage of retrieved results, which are critical for identifying specific events in large video datasets.

\medskip 
\noindent \textbf{Implementation Details} \quad For the RAPID baseline system, as outlined in Subsection \ref{sec:maths}, we employed pre-trained models from the Sentence Transformers framework\footnote{\url{https://sbert.net/}} for the \textsf{Multi-modal Embedding Model}. Specifically, we experimented sequentially with two models, namely \href{https://huggingface.co/sentence-transformers/clip-ViT-L-14}{\texttt{clip-ViT-L-14}} and \href{https://huggingface.co/sentence-transformers/clip-ViT-B-32}{\texttt{clip-ViT-B-32}}, with embedding dimensions of $768$ and $512$, respectively. To perform fast vector similarity searches, we utilized the FAISS-GPU vector database \cite{8733051}. For the \textsf{Large Language Model} in the query augmentation phase, we employed the state-of-the-art GPT-4o API from OpenAI\footnote{\url{https://platform.openai.com/docs/models/gpt-4o}}.

\subsection{Evaluation Results}

\noindent \textbf{Impact of Location-Augmented Queries}\quad We compared two types of queries: \textit{naive queries}, which focus on describing the subjects and their prominent actions in the event frames as provided in our dataset, and \textit{location-augmented queries}, where the naive queries are enhanced through the RAPID drafting process by incorporating additional location-based context.

\begin{table}[!ht]
    \centering
    \caption{Performance Comparison Between Naive and Location-Augmented Queries Across Different Types of Target Frames}
    \resizebox{\textwidth}{!}{%
    \label{tab:location}
    \begin{tabular}{cllccccccc}
        \toprule
        \textbf{Model} & \textbf{Frame Type} & \textbf{Query Type} & \textbf{P@1} & \textbf{P@5} & \textbf{P@10} & \textbf{P@20} & \textbf{R@10} & \textbf{R@20} & \textbf{MRR} \\
        \midrule
        \multirow{6}{*}{CLIP-ViT-B/32} & \multirow{2}{*}{Type 1} & Naive & 0.181 & 0.131 & 0.099 & 0.667 & 0.118 & 0.155 & 0.269 \\
                                        &                         & \cellcolor{Cyan!10}\textbf{Location-augmented} & \cellcolor{Cyan!10}\textbf{0.190} & \cellcolor{Cyan!10}\textbf{0.137} & \cellcolor{Cyan!10}\textbf{0.102} & \cellcolor{Cyan!10}\textbf{0.073} & \cellcolor{Cyan!10}\textbf{0.136} & \cellcolor{Cyan!10}\textbf{0.192} & \cellcolor{Cyan!10}\textbf{0.299} \\
                                        & \multirow{2}{*}{Type 2} & Naive & 0.079 & 0.105 & 0.075 & 0.051 & 0.114 & 0.154 & 0.204 \\
                                        &                         & \cellcolor{Cyan!10}\textbf{Location-augmented} & \cellcolor{Cyan!10}\textbf{0.184} & \cellcolor{Cyan!10}\textbf{0.113} & \cellcolor{Cyan!10}\textbf{0.088} & \cellcolor{Cyan!10}\textbf{0.062} & \cellcolor{Cyan!10}\textbf{0.136} & \cellcolor{Cyan!10}\textbf{0.195} & \cellcolor{Cyan!10}\textbf{0.298} \\
                                        & \multirow{2}{*}{All}    & Naive & 0.166 & 0.127 & 0.095 & 0.064 & 0.118 & 0.155 & 0.260 \\
                                        &                         & \cellcolor{Cyan!10}\textbf{Location-augmented} & \cellcolor{Cyan!10}\textbf{0.192} & \cellcolor{Cyan!10}\textbf{0.134} & \cellcolor{Cyan!10}\textbf{0.100} & \cellcolor{Cyan!10}\textbf{0.071} & \cellcolor{Cyan!10}\textbf{0.136} & \cellcolor{Cyan!10}\textbf{0.193} & \cellcolor{Cyan!10}\textbf{0.300} \\
        \cdashline{1-10}
        \multirow{6}{*}{CLIP-ViT-L/14} & \multirow{2}{*}{Type 1} & Naive & 0.220 & 0.200 & 0.153 & 0.100 & 0.177 & 0.231 & 0.334 \\
                                        &                         & \cellcolor{Cyan!10}\textbf{Location-augmented} & \cellcolor{Cyan!10}\textbf{0.271} & \cellcolor{Cyan!10}\textbf{0.223} & \cellcolor{Cyan!10}\textbf{0.169} & \cellcolor{Cyan!10}\textbf{0.112} & \cellcolor{Cyan!10}\textbf{0.214} & \cellcolor{Cyan!10}\textbf{0.277} & \cellcolor{Cyan!10}\textbf{0.391} \\
                                        & \multirow{2}{*}{Type 2} & Naive & 0.158 & 0.150 & 0.101 & 0.070 & 0.148 & 0.209 & 0.271 \\
                                        &                         & \cellcolor{Cyan!10}\textbf{Location-augmented} & \cellcolor{Cyan!10}\textbf{0.237} & \cellcolor{Cyan!10}\textbf{0.176} & \cellcolor{Cyan!10}\textbf{0.130} & \cellcolor{Cyan!10}\textbf{0.088} & \cellcolor{Cyan!10}\textbf{0.209} & \cellcolor{Cyan!10}\textbf{0.274} & \cellcolor{Cyan!10}\textbf{0.369} \\
                                        & \multirow{2}{*}{All}    & Naive & 0.193 & 0.144 & 0.095 & 0.052 & 0.173 & 0.228 & 0.324 \\
                                        &                         & \cellcolor{Cyan!10}\textbf{Location-augmented} & \cellcolor{Cyan!10}\textbf{0.216} & \cellcolor{Cyan!10}\textbf{0.163} & \cellcolor{Cyan!10}\textbf{0.108} & \cellcolor{Cyan!10}\textbf{0.056} & \cellcolor{Cyan!10}\textbf{0.213} & \cellcolor{Cyan!10}\textbf{0.276} & \cellcolor{Cyan!10}\textbf{0.388} \\
        \bottomrule
    \end{tabular}
    }
\end{table}

The results in Table \ref{tab:location} show a significant improvement in performance across all metrics when using location-augmented queries compared to naive queries, across all types of target frames. Notably, for event frames with difficult-to-interpret contexts, RAPID's augmented queries resulted in a 36\%-46\% increase in MRR across both models, demonstrating the effectiveness of location augmentation.

Additionally, this experiment demonstrates that the \texttt{CLIP-ViT-L/14} model consistently outperforms the \texttt{CLIP-ViT-B/32} model. The larger embedding dimension of 768 in the \texttt{L/14}, compared to 512 in the \texttt{B/32}, allows it to capture more detailed information, particularly when the query is enhanced with location-based context. Therefore, when participating in the AI Challenge, we selected the \texttt{CLIP-ViT-L/14} as the multi-modal embedding model for RAPID.

\medskip
\noindent \textbf{Comparison to AIC'24 Baseline}\quad We compared the best version of the RAPID system against the baseline provided by the competition organizers, which was also used by several participating teams, using the dataset we developed.

\begin{table}[!ht]
    \centering
    \caption{MRR Comparison Between AIC'24 Baseline and RAPID Across Different Target Frame Types}
    \label{tab:mrr}
    \begin{tabular}{lccc}
        \toprule
        \textbf{Method}     & \textbf{Type 1} & \textbf{Type 2} & \textbf{All} \\
        \midrule
        AIC'24 Baseline     & 0.382           & 0.362           & 0.379       \\
        \cellcolor{Cyan!10}\textbf{RAPID}   & \cellcolor{Cyan!10}\textbf{0.392} & \cellcolor{Cyan!10}\textbf{0.368} & \cellcolor{Cyan!10}\textbf{0.388} \\
        \bottomrule
    \end{tabular}
\end{table}

The results in Table \ref{tab:mrr} indicate that RAPID shows a slight improvement over the baseline provided by the competition organizers. While the baseline is robust, RAPID demonstrates marginal gains in handling both contextually clear and less discernible target frames.

\section{Discussion, Limitations, and Future Works}

In this paper, we introduced RAPID, a novel approach designed to enhance the precision of text-based video event retrieval by leveraging LLMs to provide relevant background details and location-based context. This method effectively bridges the semantic gap between users' textual input and the complex visual content of video scenes. Our experiments demonstrated that RAPID achieves competitive results, improving the system’s ability to capture nuanced contextual information and deliver more accurate retrieval outcomes. By utilizing parallel processing, RAPID generates multiple relevant background queries simultaneously, enriching contextual understanding across diverse video scenarios and improving system robustness. Furthermore, augmenting textual input queries significantly optimizes performance by fully utilizing the multi-modal capabilities of the CLIP-based model and the structured representation of video frames, leading to substantial improvements in both retrieval accuracy and contextual relevance.

Despite the promising results, several limitations must be acknowledged. One significant challenge arises when the model encounters scenes with unexpected or uncommon content that LLMs have not been extensively trained on, which may lead to incorrect predictions or suggestions, adversely affecting the accuracy of retrieval outcomes. Additionally, environmental conditions in the keyframes, such as varying light intensity or inadequate background representations, can interfere with scene clarity, leading to irrelevant or misleading contextual information. This reduces the precision of the retrieval process and highlights the need for more robust model generalization across diverse scenarios.

In the future, we plan to incorporate additional inputs, such as filtering with image tags or even audio, to enhance retrieval accuracy in more complex scenarios. Additionally, fine-tuning the multi-modal embedding model offers promising potential for further improving system performance.

\bibliographystyle{ieeetr}
\bibliography{references}
%\printbibliography

\end{document}